\documentclass{vldb}

\usepackage{balance}
\usepackage{etex}
\usepackage{booktabs} % For formal tables
\usepackage{color}
\usepackage{hyperref}
\usepackage{listings}
\usepackage{multirow}
\usepackage{enumitem}
\usepackage{textcomp}
\usepackage[T1]{fontenc}
\usepackage{upquote}
\usepackage{times}

\usepackage{subcaption}
\usepackage[font=small, labelfont=bf]{caption}
\usepackage{amsmath,amssymb,amsfonts,graphicx,nicefrac,mathtools}
\usepackage{thmtools,thm-restate}
\usepackage{xspace}
\usepackage{soul}
\usepackage{array}
\usepackage{cleveref}
\usepackage{makecell}
\usepackage[long]{optidef}

\usepackage{tikz}
\usepackage{pgf}
\usetikzlibrary{positioning,shapes,shadows,arrows,snakes,automata,backgrounds,petri}

\usepackage{syntax}

\usepackage[linesnumbered,ruled,resetcount]{algorithm2e}

\input{vldbreffix.tex}

\xspaceaddexceptions{[]\}(}

    {\end{list}}

\newcommand{\stitle}[1]{\vspace{0.25em}\noindent\textbf{#1}}

\newcommand{\helix}{{\sc Helix}\xspace}

\newcommand{\name}{\helix}

\crefname{example}{Example}{Examples}

\crefname{theorem}{Theorem}{Theorems}

\crefname{lemma}{Lemma}{Lemmas}

\crefname{corollary}{Corollary}{Corollaries}

\crefname{constraint}{Constraint}{Constraints}

\crefname{invariant}{Invariant}{Invariants}

\crefname{definition}{Definition}{Definitions}

\crefname{problem}{Problem}{Problems}

\DeclareMathOperator*{\argmin}{argmin}

\vldbTitle{Helix: Accelerating Human-in-the-loop Machine Learning}
\vldbAuthors{Doris Xin,  Litian Ma, Jialin Liu, Stephen Macke, Shuchen Song, Aditya Parameswaran}
\vldbDOI{https://doi.org/10.14778/3229863.3236234}

\begin{document}
\pagenumbering{gobble}

\title{ {\huge \name}: Accelerating Human-in-the-loop Machine Learning}
% Backronyms in case we ever need it: Holistic End-to-end Learning and Information eXtraction

%%%%%%%%%      Authors       %%%%%%%%%%%%%%%%%%%%
\numberofauthors{1} 
%%%%%%%%%      End of Authors       %%%%%%%%%%%%%

\author{Doris Xin,  Litian Ma, Jialin Liu, Stephen Macke, Shuchen Song, Aditya Parameswaran \\
\affaddr{University of Illinois (UIUC)} \\
\affaddr{\{dorx0,litianm2,jialin2,smacke,ssong18,adityagp\}}@illinois.edu\;}

\maketitle

\begin{abstract}
%!TEX root=helix.tex

Data application developers and data scientists spend an inordinate amount of time
{\em iterating} on machine learning (ML) work\-flows---by modifying
the data pre-processing, model training,
and post-processing steps---via trial-and-error
to achieve the desired model performance.
Existing work on accelerating machine learning focuses on speeding up
{\em one-shot execution} of workflows, 
failing to address the incremental and dynamic
nature of typical ML development.
We propose \name, a declarative machine learning system
that accelerates iterative development 
by optimizing workflow execution end-to-end and across iterations.
\name minimizes the runtime per iteration
via program analysis and intelligent reuse of previous results,
which are selectively materialized---trading off the cost of materialization for potential future benefits---to speed up future iterations.
Additionally, 
\name  offers a graphical interface to visualize workflow DAGs and compare versions to facilitate iterative development.
Through two ML applications, 
in classification and in structured prediction,
attendees will experience the succinctness of \name's programming interface and
the speed and ease of iterative development using \name. 
In our evaluations, 
\name achieved up to an order of magnitude reduction in cumulative run time compared to state-of-the-art
machine learning tools.

\end{abstract}

\section{Introduction}
\label{sec:intro}
%!TEX root=helix.tex
Development of real-world machine learning applications 
typically begins with a simple workflow,
which evolves over time 
as application developers iterate on it to improve performance.
Using existing tools, every single small change
to the workflow results in several hours of recomputation from scratch,
even though the change may only affect a small portion
of the workflow.
For example, changing the regularization
parameter should only retrain the model but not rerun data pre-processing.
One approach to mitigate this expensive recomputation
is to materialize every single intermediate
that does not change across iterations, but this approach requires
programming overhead to 
keep track of changes across iterations,
as well as to deal with how and when to materialize intermediates, 
and to reuse them in subsequent iterations.
Since this is so cumbersome, developers often 
opt to instead rerun the entire workflow from scratch.

Unfortunately, existing machine learning systems fail to provide robust support for rapid
\textit{iteration} on machine learning workflows. 
For example, KeystoneML~\cite{sparks2017keystoneml} aims at optimizing
the one-shot execution of workflows
by applying techniques such as
common subexpression elimination and caching.
Columbus~\cite{Zhang2016Columbus}
focuses on optimizing multiple feature selection steps within one iteration.
DeepDive~\cite{zhang2015deepdive},
targeted at knowledge-base construction, 
materializes the results of all feature extraction
and engineering steps.
While this na{\"i}ve materialization approach 
speeds up iterative development in certain settings, 
it can be wasteful and time-consuming.

We demonstrate \name, {\em a declarative, general-purpose end-to-end machine learning system 
that accelerates iterative machine learning application development} with three key features:

\stitle{Declarative domain specific language.} 
{\em Data scientists write code in a simple, intuitive, and modular domain-specific language (DSL) built on Scala}, {\em while also employing UDFs as needed} for imperative code, say for feature extraction or transformation. 
This interoperability allows data scientists to leverage existing functions and libraries 
on JVM and Spark-specific operators. 

\stitle{Iterative execution optimization.} 
\name represents the machine learning workflow programmed in our DSL
as a directed acyclic graph (DAG) of data collections. 
For each node (representing an intermediate result), 
\name decides whether to materialize it by considering the maximum storage budget, 
the time to compute the node and all of its ancestors, 
and the size of the output---this is the {\em materialization} problem.
Then, during subsequent iterations, \name determines whether to read the result for a node from persistent storage (if previously materialized), or to compute it from the input---this is the {\em recomputation} problem. 
We found that recomputation is
in {\sc PTIME} via a non-trivial reduction to {\sc Max-Flow} using the
{\sc Project Selection Problem}~\cite{kleinberg2006algorithm}, 
while materialization is {\sc NP-Hard} via a reduction from the {\sc Knapsack Problem}.
We propose a simple cost model used in an online algorithm 
to provide an approximate solution to the materialization problem.
Figure~\ref{fig:perf} (described later) shows that \name provides 60\% to an order of magnitude reduction in 
cumulative run time reduction compared to state-of-the-art
tools like DeepDive and KeystoneML.

\stitle{Workflow versioning and visualization tool.} 
We build a versioning and visualization tool on top of \name, enabling the management of workflow versions, 
execution plan visualization, and version comparison. 
Users can easily track the evolution of a workflow, 
including the changes to hyperparameters, 
feature selection, and the performance impact
of each modification on the workflow.

\vspace{2pt}
\noindent 
Our demonstration aims to: 
a) Highlight \name's succinct yet flexible declarative DSL for programming end-to-end machine learning workflows. 
b) Demonstrate how \name accelerates iterative machine learning application development by providing 
  (1) End-to-end optimization of the entire workflow; 
  (2) Automatic detection of the operator changes; 
  (3) Intelligent materialization of intermediate results for maximizing reuse in subsequent executions. 
c) Show how \name's graphical interface can support debugging and result analysis during workflow development.
The attendees will be able to interact with the \name system 
via a graphical interface that includes four main modules: 
code editor, 
workflow DAG visualization tool (shown side-by-side with the code), 
workflow versions browser, and workflow version comparison tool. 
The DAG visualization tool helps users 
explore optimizations 
to the execution plan. 
The version browser and comparison tool 
allows users to gain insights 
into relationships among features, models, 
and performance metrics, 
thus providing developers effective guides on how to 
fine-tune the model to save exploration time. 

Note that the techniques and abstractions
involved in building \name are 
{\em general}---wrappers for
other ML and data processing frameworks
can be easily implemented
while using the same core optimization engines 
and programming model.

\section{System Overview}
\label{sec:overview}
%!TEX root=helix.tex

\begin{figure*}
\centering
	\includegraphics[width=0.85\textwidth]{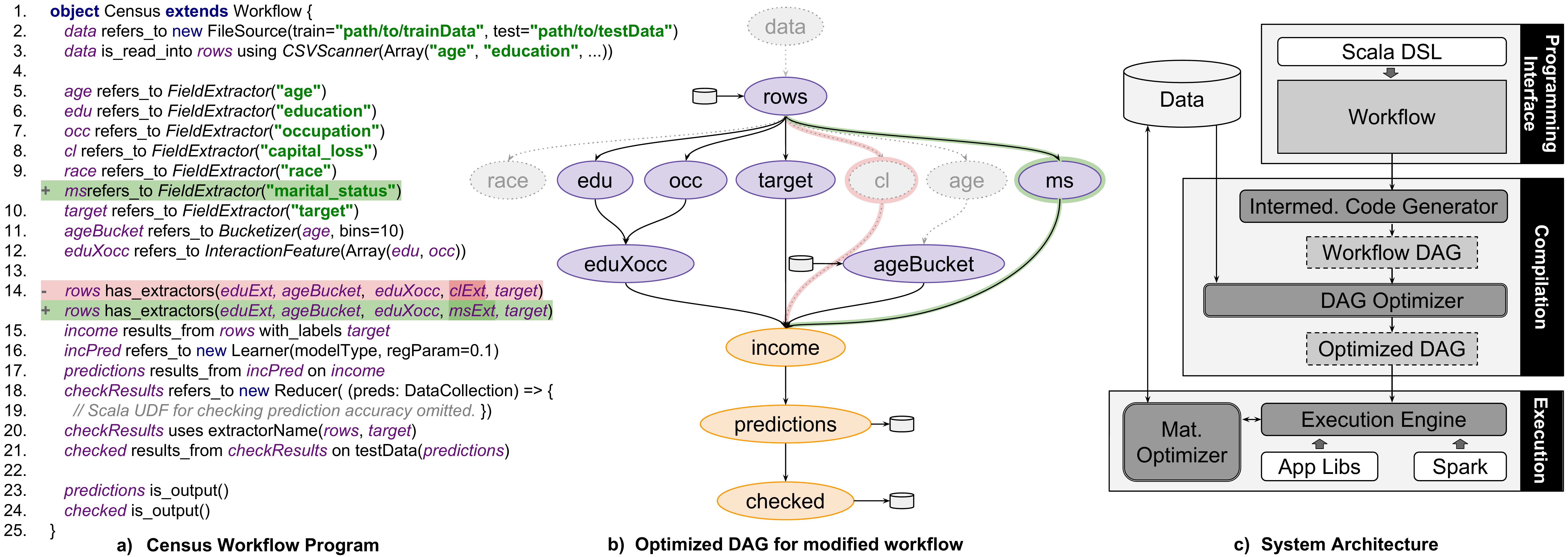}
	\vspace{2pt}
	\caption{a) Example workflow in \name DSL for the Census application, 
	with $+$/$-$ indicating iterative changes. 
	b) Optimized execution plan for the modified workflow in a).
	Operators for data pre-processing are in purple, and machine learning in orange; 
	operators from a) pruned at execution time are grayed out.
	Nodes highlighted in red and green correspond to the code changes in a).
	c) System architecture.
	}
	\label{fig:dag}
\end{figure*}

The \name backend
comprises a domain specific language (DSL) in Scala  as the
programming interface, a compiler for the DSL, and an execution engine.
\Cref{fig:dag}c) illustrates the backend architecture.
The compiler first translates the program written in the DSL 
into a DAG of intermediate results (associated with the corresponding operators
that generated them),
which is then optimized to minimize overall execution time, 
by pruning extraneous operations (or equivalently, intermediate results), 
reordering operations, 
and reusing results from previous iterations when applicable.
The execution engine uses an online algorithm that
determines at runtime the set of intermediate results to materialize
in order to minimize execution time for subsequent iterations.
We provide a brief overview of each of these three components below.

\subsection{Programming Interface}
\name's DSL is akin to KeystoneML's DSL
for constructing ML pipelines,
with the added benefits of user-friendly 
data structures for data pre-processing.
\helix users program their entire procedure
in a single Scala interface called Workflow. 
Users can directly embed Scala expressions 
as user-defined functions (UDFs)
into declarative statements in the DSL, 
in the same fashion that SparkSQL
supports inline SQL UDF registration~\cite{armbrust2015sparksql}. 
Figure~\ref{fig:dag}a) 
shows an example workflow in the \name DSL 
for the Census application 
that will be described in Section~\ref{sec:apps}.
The DSL facilitates 
elaborate data pre-processing 
and complex machine learning (ML) model development 
with the following features.
With a handful of operator types,
the DSL supports 
both fine-grained and coarse-grained feature engineering, 
as well as both supervised and unsupervised learning.
The DSL has been used to implement workflows 
in {\em social sciences, information extraction, computer vision, and natural sciences,}
all under 100 lines of code per workflow.
Users can easily extend the default set of operators to adapt to their custom use cases
by providing only the UDF without copying boilerplate code.
\name's data structure for  pre-processing 
maintains features in human-readable format 
for ease of development
and automatically 
converts it into a compatible format for ML.

\subsection{Compiler}
During the compilation phase,  high-level DSL
declarations in a Workflow are first translated 
into a DAG
of operations (or equivalently, intermediate results)  
using the \textit{intermediate code generator}.
Figure~\ref{fig:dag}b) shows an example of the operations DAG
compiled from the program in Figure~\ref{fig:dag}a).
The \textit{DAG optimizer} analyzes the generated DAG
along with relevant data, including the input data and any materialization
of intermediate results from previous executions,
to produce a \textit{physical execution plan}, with the optimization objective
of \textit{minimizing the latency of the current iteration.}
This  involves several components:

\stitle{Iterative change tracker.}
To avoid the inefficiencies of rerunning invariant operations, 
\name automatically detects changes to an operator from the last iteration
and invalidates all results affected by the changes via dependency analysis.
Unfortunately, the problem of determining operator equivalence for arbitrary functions 
is undecidable as per Rice's Theorem~\cite{rice1953classes},
with extensive bodies of work in the
programming language community dedicated to 
solving it for specific classes of programs. 
Currently, \name supports change detection 
via source code version control;
covering more general cases is future work.
Figure~\ref{fig:dag}a) shows highlighted 
changes automatically detected by \name 
between two versions of a workflow
($+$/$-$ indicates statements that are added/deleted).

\stitle{Program slicing component.}
\name applies program slicing techniques from compilers 
to prune extraneous operations 
that do not contribute to the final results. 
Feature selection is a prevalent practice in machine learning, 
and this component uses fine-grained data provenance to
automatically eliminate computation for features that do not impact the model,
without any code change by the user.

\stitle{Recomputation component.}
The DAG optimizer in the compiler determines 
the optimal reuse policies that minimize execution 
time of the current iteration 
given results from previous iterations.
Formally, let $G = (N, E)$ be a DAG of operations.
Each $n_i \in N$ has a {\em compute cost} $c_i$
and a {\em load cost} $l_i$.
Additionally, each node is assigned a state from $S = \{load, compute, prune\}$,
with the {\em prune constraint} that stipulates 
that a node in $compute$ cannot have parents in $prune$ 
(i.e., the parents of a node must be available for that node to be computed).
Let $s : N \rightarrow S$ be the state assignments
and $\mathbb{I}$ be the indicator function.
The objective of the recomputation problem is finding $s$:
\begin{equation}
\argmin\limits_{s} \sum\limits_{n_i \in N} \mathbb{I}\{s(n_i) = compute \}c_i + \mathbb{I}\{s(n_i) = load \} l_i
\vspace{-5pt}
\end{equation}
This cannot be solved via a simple traversal of the DAG
due to the prune constraint. 
While loading a node $n_i$ allows us to prune all of its ancestors $A(n_i)$,
the actual run time reduction incurred by loading $n_i$ 
depends on the states of all descendants of each $n_j \in A(n_i)$.
For example, if $l_k \gg c_k$ for a node $n_k$ that is a child of some $n_j \in A(n_i)$,
the run time is minimized by keeping $n_j$ and computing $n_k$ from it.
We prove that this problem is polynomial-time reducible to the {\sc Project Selection problem} \cite{kleinberg2006algorithm}, 
a variant of {\sc Max-Flow},
and devise an efficient {\sc PTIME} algorithm to compute the optimal 
plan via this reduction\cite{dorx2017}.

\vspace{2pt}
\noindent Figure~\ref{fig:dag}b) shows an example optimized plan. 
Each node corresponds to the result 
of an operator declared in Figure~\ref{fig:dag}a),
with operators for data pre-processing in purple and machine learning in orange.
Nodes with a drum to the left are reloaded from disk, 
whereas nodes with a drum to the right are materialized. 
Operators in the source code that are pruned during execution are grayed out.
Iterative changes to the code are highlighted in red and green in~\ref{fig:dag}a).

%!TEX root=helix.tex

\subsection{Execution Engine}
The execution engine
executes the physical plan produced by the compiler, using 
Spark~\cite{zaharia2012resilient} as the main
backend for data processing, supplemented with
JVM libraries 
for application-specific needs.
 
During execution, the {\em materialization optimizer}
chooses intermediate results to persist 
(with a maximum storage constraint)
in order to {\em minimize the latency of future iterations}, 
using runtime statistics
from the current and prior executions for guidance.
This optimization problem is complicated by two practical challenges: 
1) the total number of iterations the user will perform is not known a-priori;
2) changes to the workflow in future iterations are unpredictable, 
making it difficult to determine the intermediate results 
that can be reused in iterations that follow.
Even in the simplest case, with the strong assumption that 
the user will carry out only one more iteration,
and all results from the current iteration will be reusable in the next,
it can be shown, via a reduction from the {\sc Knapsack Problem},
that this optimization problem is still {\sc NP-Hard}.
Additionally, the decision to materialize must take place 
immediately upon operation completion, 
as it is prohibitive to cache multiple intermediate results 
for deferred decisions.
Thus, an online algorithm 
is needed to make decisions quickly in real-time.
We use a simple cost model to determine 
the set of intermediate results to materialize 
as they become available.
Recall an operator $n_i$ is associated with a load cost $l_i$ and a compute cost $c_i$.
At iteration $t$, the reduction in execution time at iteration $t+1$, 
from materializing $n_i$ at $t$, can be approximated as
$r_i = 2l_i - ( c_i + \sum_{n_j \in A(n_i)} c_j )$.
If $r_i$ is negative and the data size for $n_i$ 
is less than the remaining storage budget, then materialize $n_i$.
Although this model ignores the dependencies between other operators and $A(n_i)$, it is cheap to compute and effective in practice, while satisfying the online constraint.
Our ongoing work investigates predicting reuse probability 
based on user studies and workflow features.

\begin{figure}[b]
\centering
\includegraphics[width=0.16\textwidth]{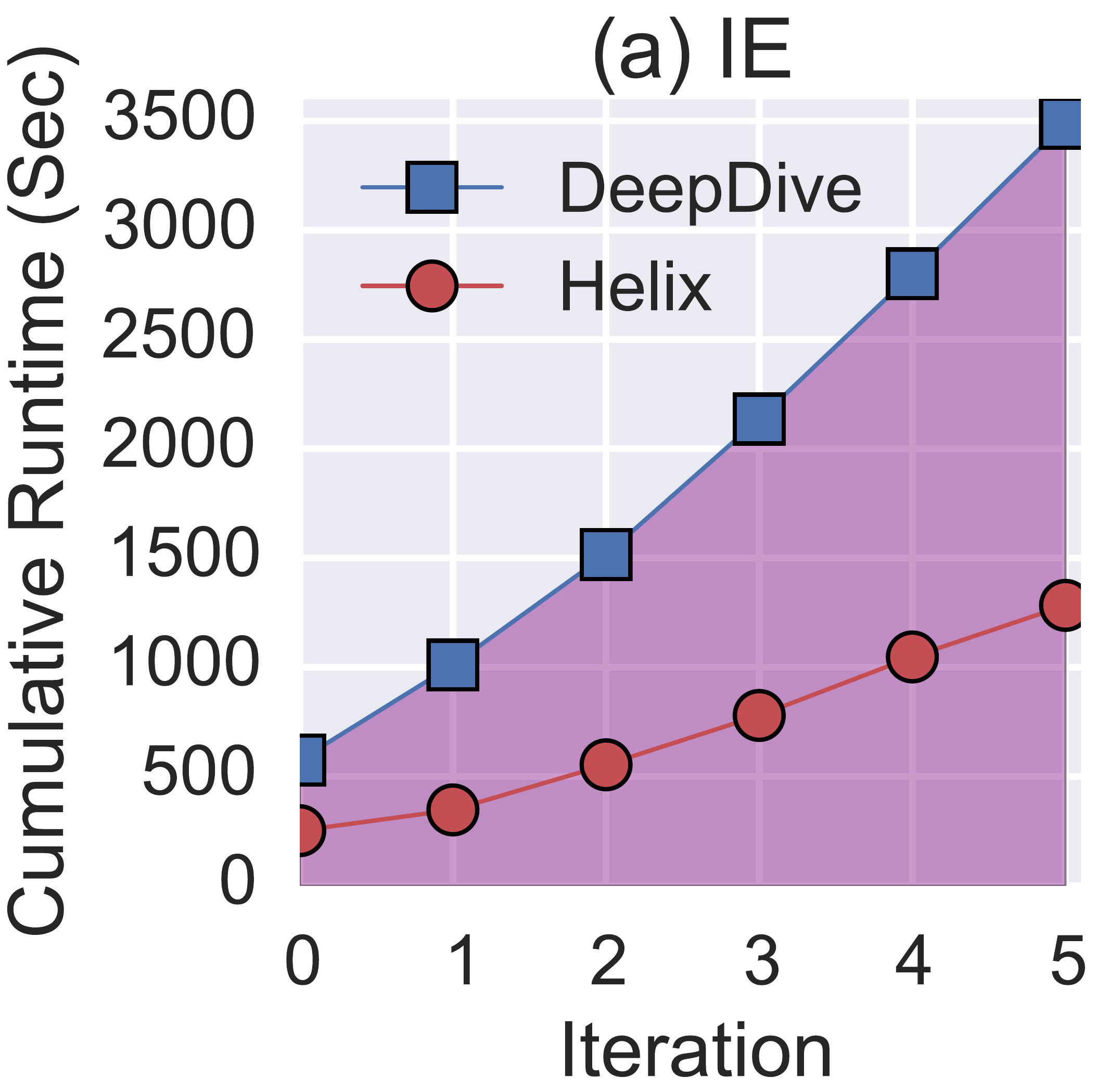}
\hspace{-5pt}
\includegraphics[width=0.26\textwidth]{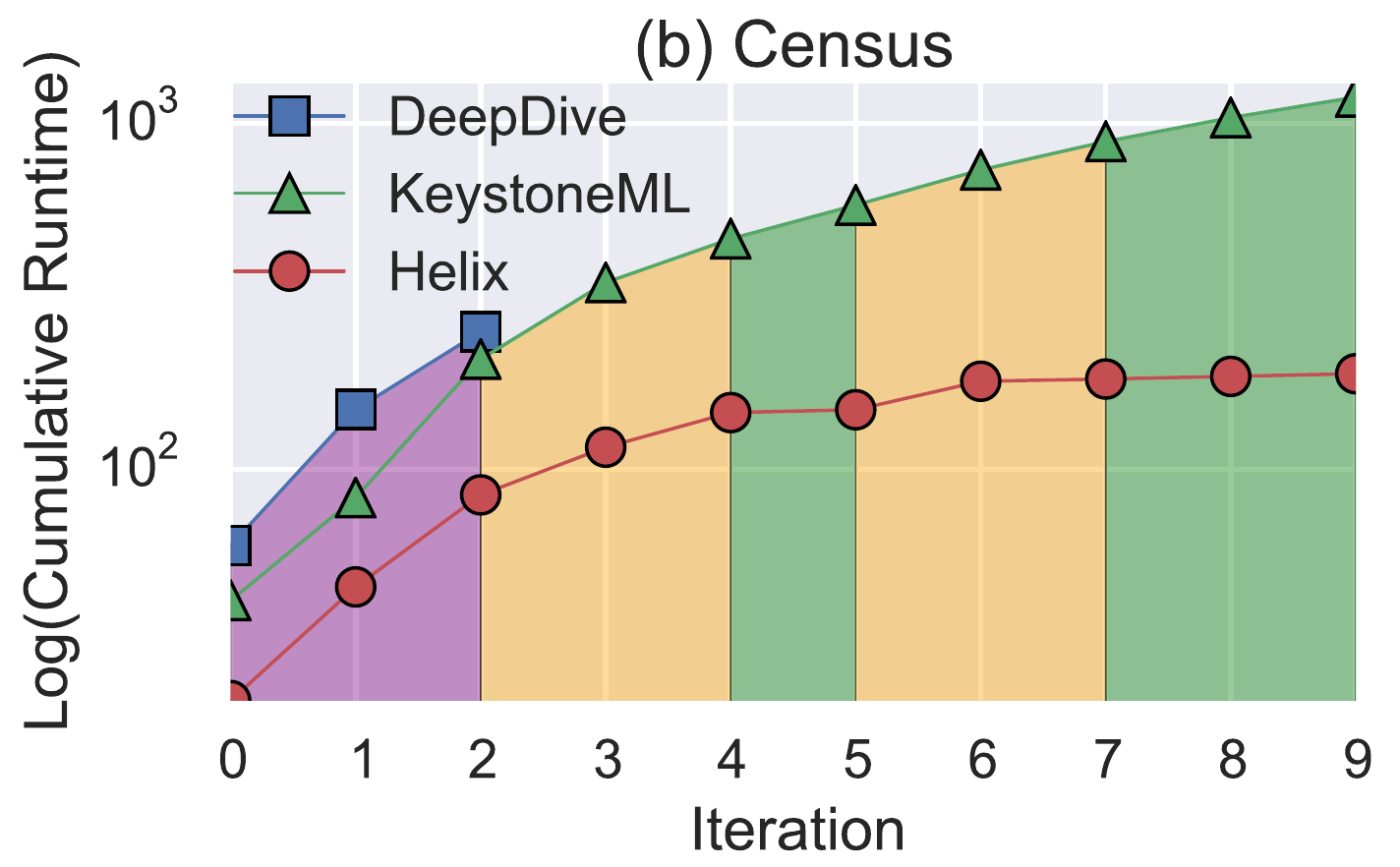}
\caption{Cumulative runtime comparison with (a) DeepDive on an IE task.
(b) DeepDive and KeystoneML on a classification task.}
\label{fig:perf}
\end{figure}

\subsection{Performance Gains}
We show preliminary experiments
comparing  \name with two similar ML systems, DeepDive~\cite{zhang2015deepdive} and KeystoneML~\cite{sparks2017keystoneml},
on the two applications to be described in Section~\ref{sec:apps}.
KeystoneML is absent in Figure~\ref{fig:perf}(a) 
because it is not equipped to handle information extraction (IE) tasks,
whereas DeepDive has missing data for iteration $>2$ in Figure~\ref{fig:perf}(b)
because its ML and evaluation components are not user-configurable.
To show the type of modification in each iteration, 
we use purple to indicate a data pre-processing change (e.g., adding a feature),
orange for ML (e.g., adding regularization), 
and green for evaluation (e.g., changing metrics).

Figure~\ref{fig:perf}(a) shows that for the IE task, 
\name 's cumulative run time is {\em 60\% lower than that of DeepDive},
due to judicious materialization of only intermediates that help with future iterations, in contrast to DeepDive's materialize-all approach.
Figure~\ref{fig:perf}(b) shows all three systems' performance on a classification task, 
where \name shows nearly {\em an order of magnitude reduction} in cumulative run time. 
Note that in post processing iterations (green), 
\name has near zero runtimes 
due to high reuse rate. 
ML iterations (orange) has slightly higher runtime 
but less than data pre-processing iterations (purple), 
which have the least amount of reuse.
For a never-materialize system such as KeystoneML,
the rerun time is constantly large regardless of what has been changed.

\section{Demonstration Description}
\label{sec:demo}
%!TEX root=helix.tex

\begin{figure*}[t]
\centering
\includegraphics[width=0.85\textwidth]{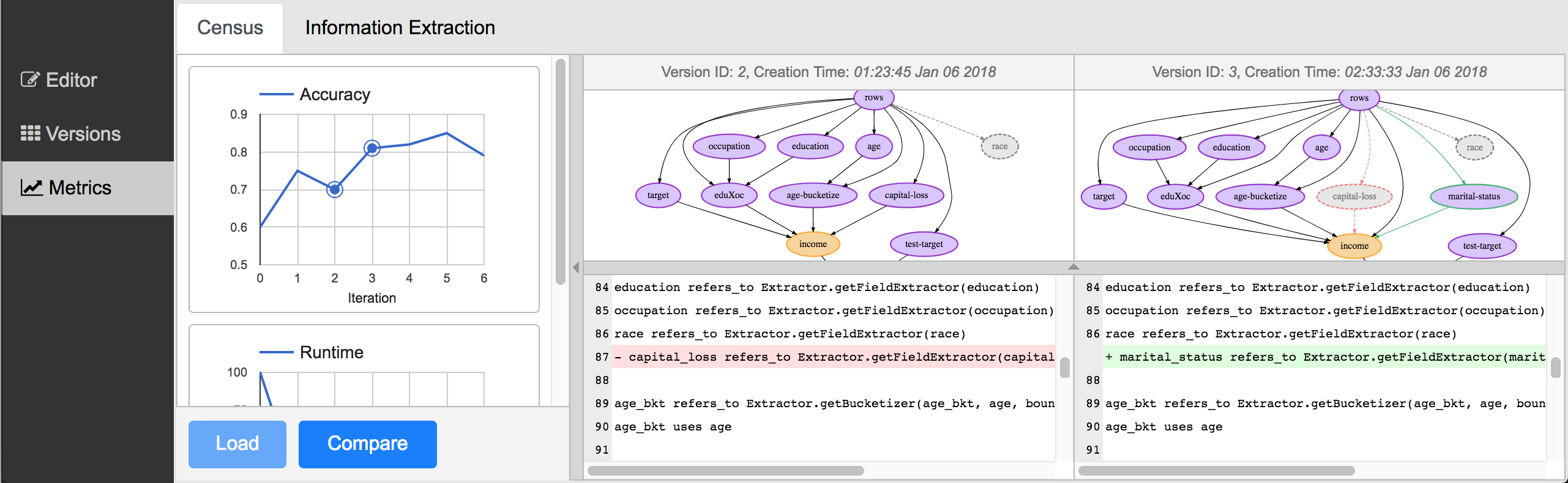}
\vspace{2pt}
\caption{Metrics aggregation and version comparison.
Users can select and compare specific versions, 
represented as points on the metric trend lines,
for code change and visualized execution plans,
in order to better understand the performance impact of specific modifications.
}
\label{fig:metrics}
\end{figure*}

We will demonstrate the ease and speed of iterative machine learning development using \name
through two distinct ML applications.
We compare with the unoptimized version of \name to help attendees 
appreciate the gains of \name's optimizations.

\label{sec:apps}
\stitle{Applications.} 
\textit{1) Census:}
This application illustrates a simple classification task 
with straightforward features from structured input. 
The dataset from \cite{Lichman:2013}
contains demographic information, such as age, education, occupation, 
used to predict whether a person's annual income is >50K.
The complexity of this application is representative of applications 
from the social and natural sciences, where well-defined variables
are being studied for covariate analysis.
Code for this workflow is shown in Figure~\ref{fig:dag}a).
\textit{2) Information Extraction:}
This is a complex structured prediction task 
that identifies person mentions 
from news articles. 
In contrast to \textit{Census}, the input to this workflow is unstructured text, 
and the objective is to extract structured information instead of simple classification.
Thus, this workflow requires more data pre-processing steps to enable learning,
mirroring the typical industry setting where extensive data ETL is necessary.

\subsection{User Interface}
Attendees will interact with the \name system through a single web application
with an IDE for programming and modules for examining results and system details. 

\stitle{IDE.} The \name IDE provides \name DSL specific autocomplete and syntax highlighting to facilitate programming. 
A ``Suggest Modifications'' button lets user request 
machine-generated edits to be shown inline 
using Github-style code change highlighting, 
as illustrated in Figure~\ref{fig:dag}a),
thus allowing users to iterate rapidly on the workflow without mastering the DSL.
Once the workflow is executed, the user will be able to inspect the optimized execution plan in the DAG format,
as shown in Figure~\ref{fig:dag}b).
Individual runtime and storage for each operation are displayed by hovering over them.

\stitle{Versions.}
Users can quickly browse through all past versions 
of a workflow in a summarized view
with similar aesthetics to code version control tools such as git.
Each version is shown as a commit log entry, 
with buttons that allow users to instantly \textit{checkout} 
the code or obtain additional metadata. 
We also provide shortcuts to the version 
with the best evaluation metrics as well as 
the latest version at the top of the page.

\stitle{Metrics.}
As shown in Figure~\ref{fig:metrics}, 
the Metrics tab aggregates the evaluation 
metrics for the workflow across iterations 
into plots with the metric value 
on the y-axis and the iteration number on the x-axis.
Each point in the plot represents a version of the workflow. 
Users can select a single point to load the associated code version
or two points for comparison. 
In Figure~\ref{fig:metrics}, Version 2 and 3 are selected 
in the Accuracy plot for comparison.
The comparative view visualizes the DAG 
and highlights changes in the DAG using git-like visual comparison cues, 
in addition to showing the two versions 
of the workflow code also with changes highlighted.
This feature enables rapid exploration of the 
relationships between various metrics and 
changes to specific components of the workflow.
Understanding the impact of each past iteration 
is crucial for making effective future improvements, 
thus reducing the overall number of iterations to achieve the desired outcome.

\subsection{Guided Interaction}
Once acquainted with the system and applications, 
attendees are invited to select an application 
and execute its initial version.
Upon completion, we will describe each component 
of the UI using the initial results.
Attendees are then invited to modify 
the workflow in the IDE to optimize for 
either the prediction accuracy 
or overall runtime to achieve a certain level of accuracy.
They can also use machine-generated suggested edits to the workflow
for quick exploration without learning the DSL.
The version browser can be used 
to see past changes 
or roll back to a past version and branch out in another direction.
After the first modified version is executed, 
we will compare the execution plan with the one from the previous iteration  
to showcase the workflow change detection feature,
which allows \name to automatically reuse results for operators not affected by the changes.

Attendees can investigate the relationship 
between past versions and metric values using the Metrics tab 
to inform decisions on what to try next.
To emphasize the benefits of \name's optimizations, 
we will execute the same version twice, once with and once without optimizations,
and compare the runtimes and execution plans.
At the end of the session, attendees are invited 
to review a summary of their interactions via the Versions and Metrics tabs 
to gain a better appreciation for the progress they made in a short period of time.

\section{Related Work}
\label{sec:rel}
%!TEX root=helix.tex

A great deal of work focuses on development of end-to-end systems
for common ML operations, focusing on expressiveness for ML tasks
at the language level~\cite{zhang2015deepdive,kraska2013mlbase}
or API level~\cite{meng2016mllib,manning2014stanford}
and provides first-class support for tasks such as model selection~\cite{sparks2015tupaq},
workflow construction~\cite{pedregosa2011scikit}, feature selection~\cite{Zhang2016Columbus},
and feature engineering~\cite{zhang2015deepdive}.
Such systems typically optimize the runtime of ML pipelines
on a single node~\cite{feng2012bismarck} or in a distributed
setting~\cite{sparks2017keystoneml,armbrust2015sparksql,kraska2013mlbase}.
Another common theme is the specification of machine learning tasks through
an expressive and easy-to-use declarative programming
model~\cite{zhang2015deepdive,kraska2013mlbase}.
\helix shares some characteristics with these systems in that
it adopts many of the same goals as secondary considerations,
but is unique in that it identifies {\em iterative development} as a primary concern
and is the first system to implement novel, principled
solutions for this particular focus.

\balance

\smallskip
{\small
\noindent {\bf Acknowledgments.} We thank the anonymous reviewers for their valuable feedback. We acknowledge support from grants
IIS-1513407, IIS-1633755, IIS-1652750, and IIS-1733878 awarded by the National Science Foundation, and funds from Microsoft, 3M, Adobe, Toyota Research Institute, Google, and the Siebel Energy Institute. Doris Xin and Stephen Macke were supported by National Science Foundation Graduate Research Fellowship grants NSF DGE-1144245 and NSF DGE-1746047. The content is solely the responsibility of the authors and does not necessarily represent the official views of the funding agencies and organizations.
}

{\scriptsize
\bibliographystyle{abbrv} % vldb
\bibliography{sigproc.bib}
}

\end{document}